\documentclass{svproc}
\usepackage{url}

\usepackage{graphicx}
\usepackage{tabularx}
\usepackage{booktabs}
\usepackage{amsmath}
\usepackage{pifont}
\usepackage{amssymb}
\usepackage{scalerel,xparse}
\newcommand{\cmark}{\ding{51}}%
\newcommand{\xmark}{\ding{55}}
\usepackage[lofdepth,lotdepth]{subfig}
\NewDocumentCommand\emojismiley{}{
    \includegraphics[scale=1.0]{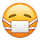}
}

\begin{document}
\mainmatter              
\title{Transformer based Automatic COVID-19 Fake News Detection System }
\titlerunning{Fake News Detection}  
%
\author{Sunil Gundapu \and Radhika Mamidi}
\authorrunning{Gundapu and Mamidi} 
%
\tocauthor{Gundapu and Mamidi}
\institute{
International Institute of Information Technology, Hyderabad\\
\email{sunil.g@research.iiit.ac.in, radhika.mamidi@iiit.ac.in}
}

\maketitle              

\begin{abstract}
Recent rapid technological advancements in online social networks such as Twitter have led to a great incline in spreading false information and fake news. Misinformation is especially prevalent in the ongoing coronavirus disease (COVID-19) pandemic, leading to individuals accepting bogus and potentially deleterious claims and articles. Quick detection of fake news can reduce the spread of panic and confusion among the public. For our analysis in this paper, we report a methodology to analyze the reliability of information shared on social media pertaining to the COVID-19 pandemic. Our best approach is based on an ensemble of three transformer models (BERT, ALBERT, and XLNET) to detecting fake news. This model was trained and evaluated in the context of the ConstraintAI 2021 shared task ``COVID19 Fake News Detection in English" \cite{gupta:kumari}. Our system obtained 0.9855 f1-score on testset and ranked 5th among 160 teams. 
\end{abstract}

\section{Introduction}

The COVID-19 pandemic is considered the global public health crisis of the whole world and the biggest problem people faced after World War II. COVID-19, a contagious disease caused by a coronavirus, has caused more than 75 million confirmed cases and 1.7 million deaths across the world till 2020 December\footnote{https://www.worldometers.info/coronavirus/}. Unfortunately, the misinformation about COVID-19 has encouraged the growing of the disease and chaos among people. During the Munich Security Council held on February 15, 2020, World Health Organization (WHO) Director-General, Tedros Adhanom Ghebreyesus \cite{datta:yadhav} stated that the world was in a war to fight not only a pandemic, but also an infodemic. So we should address the challenge of fake news detection to stop the spreading of COVID-19 misinformation.

Since the global pandemic impacts the people, there is a broader public finding information about the COVID-19, whose safety is intimidated by adversarial agents invested in spreading fake news for economic and political reasons. Besides, due to medical and public health issues, it is also hard to be totally valid and factual, leading to differences that worsen with fake news. This difficulty is compounded by the quick advancement of knowledge about the disease. As researchers gain more knowledge about the virus, claims that looked right may turn out to be false, and vice versa. Detecting this spread of COVID-19 associated fake news, thus, has become a pivotal problem, gaining notable attention from government and global health organizations (WHO, 2020), online social networks (TechCrunch, 2020), and news organizations (BBC, 2020; CNN, 2020; New York Times, 2020).

In response to the present disinformation, this paper looks at developing an efficient fake news detection architecture with respect to COVID-19. Initially, we started with developing machine learning (ML) algorithms with Term Frequency and Inverse Document Frequency (TF-IDF) feature vectors to detect misinformation on the provided dataset. These supervised TF-IDF methods are still relevant for many classification tasks and performed pretty well for fake news detection. We developed an effective ensemble model integrated with three transformer models for detecting fake news on the social media platforms. This resulted in higher accuracy and a more generalized model.

The rest of this paper is organized as follows, Section II presents some prior works related to fake news, and its spread, on social media platforms. In Section III, we describe the dataset provided in the Constraint AI- 2021 shared task. Section IV presents implemented models and framework for misinformation detection. Section V provides the discussions on the results. Finally we conclude this paper in Section VI.

\section{Related Work}

\textbf{Fake News Detection: } Fake news can be defined as inaccurate and misleading information that is growing knowingly or unknowingly \cite{chen:sin}. Recognizing the spread of false information such as rumors, fake news, propaganda, hoaxes, spear phishing, and conspiracy theories is an essential task for natural language processing \cite{throne:vla}. Gartner's \cite{tir:car} research studies explained that most people in advanced economies would believe more fake information than truthful information by 2022.

To date, so many automated misinformation detection architectures have been developed.  Rohit et al. \cite{kali:sing} provided an extensive survey to detect fake news on various online social networks. Ghorbani et al. \cite{zhang:ghori} presented an inclusive overview of the recent studies related to misinformation. Furthermore, they described the impact of misleading information, shown state-of-the-art fake news detection systems, and explored the disinformation detection datasets. The majority of the fake news detection models developed using supervised machine learning algorithms to classify the data as misleading or not \cite{khan:khon}. This supervised classification is concluded by comparing the user input text with some already created corpora containing genuine and misleading information \cite{elha:li}. 

Aswini et al. \cite{thota:thilak} proposed a deep learning architecture with various word embeddings for Fake News Challenge (FCN-1) dataset\footnote{http://www.fakenewschallenge.org/}. They developed the architecture to accurately predict the stance between a given pair of news headlines and the corresponding article/body. On the same FCN-1 dataset, Sean et al. \cite{sean:doug} developed an average weighted model of TalosCNN and TalosTree called TalosComb. TalosCNN is a convolutional neural network with pre-trained word2vec embeddings, and TalosTree is a gradient-boosted decision tree model with SVD, word count, TF-IDF. By analyzing the relationship between the news headline and the corresponding article, Heejung et al. \cite{jwa:oh} designed the Bidirectional Encoder Representations from Transformers model (BERT) to detect misleading news articles.

\subsubsection{COVID-19:}In the case of COVID-19 fake news, a large number of misleading contents remain online on social media platforms. NLP researchers have been working on developing algorithms for the detection of online COVID-19 related disinformation. To develop any algorithm, we require a corpus. So members of the NLP community created the various fake news datasets: FakeCovid \cite{shah:nand}, ReCOVery \cite{zhou:mulay}, CoAID \cite{cui:lee}, and CMU-MisCOV19 \cite{memon:car}. Yichuan Li et al. \cite{li:jiang} developed multi-dimensional and multilingual MM-COVID corpora, which covers six languages. Mabrook et al. \cite{rakh:amri} created a large Twitter dataset related to COVID-19 misinformation. And authors developed an ensemble-stacking model with six machine learning algorithms on the created dataset for detecting misinformation. 

Elhadad et al. \cite{elha:lik} constructed a voting ensemble machine learning classifier for fake news detection that uses seven feature extraction techniques and ten machine learning models. Tamanna et al. \cite{hoss:robe} used the COVIDLIES dataset to detect the misinformation by retrieving the misconceptions relevant to the Twitter posts. For COVID-19 fake news detection and fact-checking, Rutvik et al. \cite{vijj:potl} proposed a two-stage transformer model. The first model retrieves the most relevant facts about COVID-19 by using a novel fact-checking algorithm, and the second model, by computing the textual entailment, verifies the level of truth. Adapting all these classical and hybrid related work techniques, we developed a COVID-19 fake news detection system in this paper.

\section{Dataset Description}

The ConstraintAI'21\footnote{https://constraint-shared-task-2021.github.io/} shared task organizers developed a COVID-19 fake news detection in English dataset \cite{patwa:sharma} containing 10,700 data points collected from various online social networks such as Twitter, Facebook, and Instagram, etc. From the total dataset, 6,420 data points are reserved for training, 2,140 data points are used for hyperparameter tuning as a part of the validation phase, and the remaining 2,140 social media posts are kept aside for testing.  Each dataset except the test set contains social media data points and their corresponding labels, either real or fake.

\begin{table}[ht]
\centering
\subfloat[Subtable 1 list of tables text][Dataset Statistics]{
\begin{tabular}{c@{\hspace{0.1in}}|@{\hspace{0.1in}} c @{\hspace{0.1in}}|@{\hspace{0.1in}}c}
\hline
\textbf{Corpus} & Real & Fake \\
\hline
\textbf{Train}      & 3360    & 3060    \\
\textbf{Valid}       & 1120     & 1020 \\
\textbf{Test}       & 1120     & 1020      \\
\hline
\end{tabular}}
\qquad
\subfloat[Subtable 2 list of tables text][Label-wise example]{
\begin{tabular}{l|c}
\hline
\textbf{Tweet} & \textbf{Label}\\
\hline
CDC Recommends Mothers Stop & fake\\ 
Breastfeeding To Boost Vaccine Efficacy &  \\  
1000 COVID-19 testing labs in India: ICMR & real \\ 
\hline
\end{tabular}}
\qquad
\caption{Fake news dataset information}
\end{table}

Table 1 shows the corpus size and label distribution, and if we observe, the labels in each dataset are all roughly balanced. Table 2 shows some examples from the COVID-19 fake news detection in the English dataset. We illustrate the most occurring word cloud of the real and fake data points after removing the stop words in Figures 1(a) and 1(b). In Figure 1(a), we can see unique words in real-labeled data points which don't often occur in Figure 1(b), like ``covid19", ``discharged", ``confirmed", ``testing", ``indiafightscorona", and ``indiawin", etc.; meanwhile, from Figure 1(b), we can find unique words frequently appearing in the fake articles, which include ``coronavirus", ``kill", ``muslim", ``hydroxychloroquine", ``china", and ``facebook post", but don't frequently appear in the true labeled data points. These frequent textual words can give important information to differentiate the true data points from fake ones.

\begin{figure}%
    \centering
    \subfloat[\centering Positive word cloud ]{{\includegraphics[width=5cm]{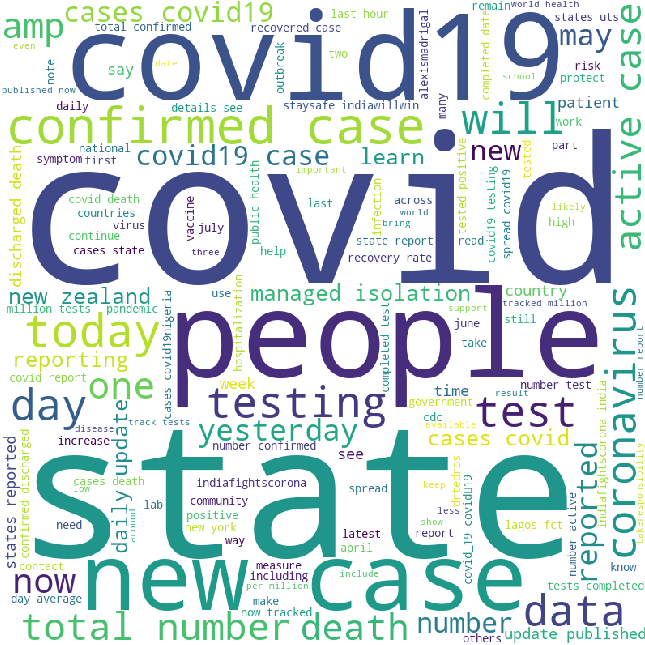} }}%
    \qquad
    \subfloat[\centering Negative word cloud ]{{\includegraphics[width=5cm]{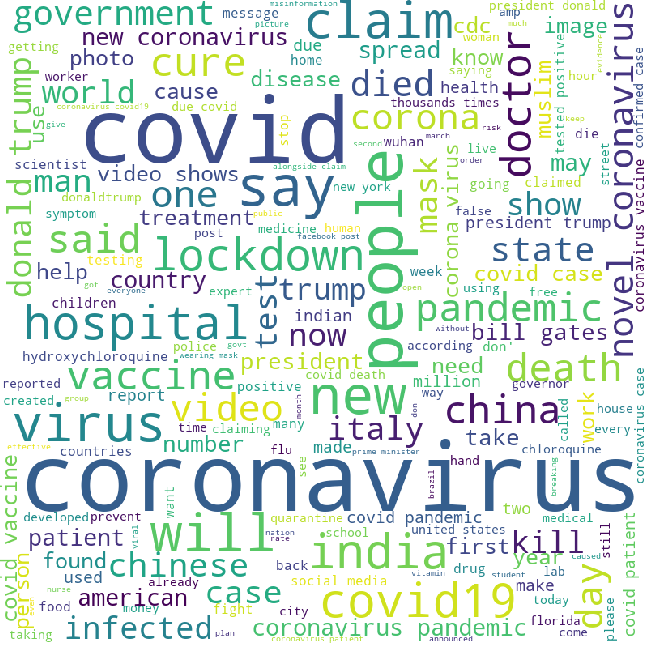} }}%
    \caption{Illustration of frequent word cloud}
    \label{fig:example}%
\end{figure}

\section{Methodology}

In this part, we present our transformer based ensemble model that is trained and tuned on the datasets which reported in the previous section. We compare our approach with various machine learning (ML) and deep learning (DL) models with different word embeddings. The full code of system architecture can be found at GitHub\footnote{https://github.com/SunilGundapu/Covid-19-fake-news-detection}.

\subsection{Data Preprocessing}

The main aim of this part is to use the NLP techniques to preprocess the input tweet data and prepare for the next step to extract the proper features. In Figure 2, we shown the detailed data preprocessing pipeline with examples.

\begin{center}
  \begin{figure}[h!]
  \makebox[\textwidth]{\includegraphics[width=12cm,height=3cm]{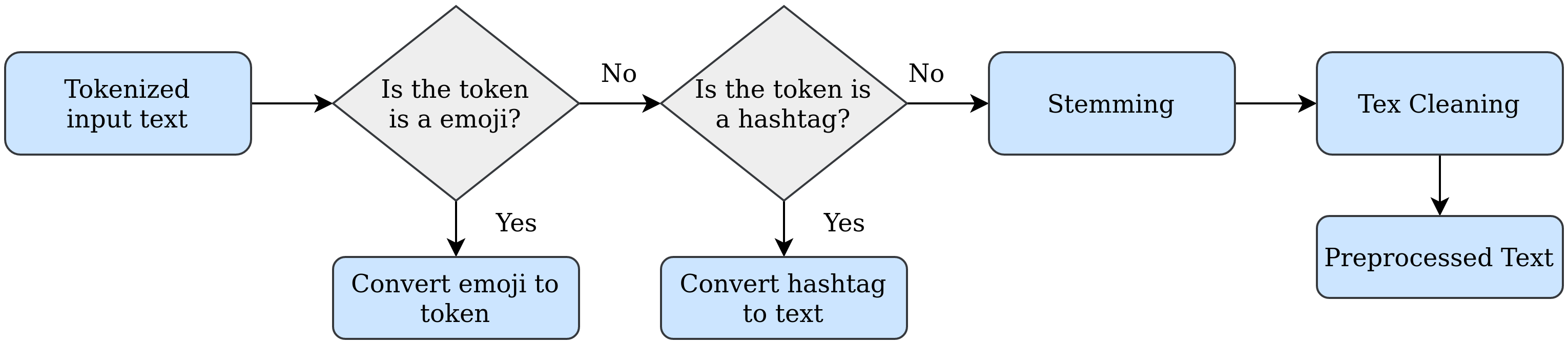}}
  \caption{Data preprocessing pipeline}
  \end{figure}
\end{center}

In the preprocessing step, we will forward the tokenized tweet through the pipeline to eliminate the noise in the fake news dataset by remove or normilize the unnecessary tokens.  The preprocessing pipeline includes the following sub-parts:

\begin{enumerate}
\item \textbf{Emoticon Conversion: } In this step, we converted the each emoticon in the tweet to text. Example: \(\emojismiley \rightarrow \text{Face with medical mask emoji}\)

\item \textbf{Handling of Hashtags: } We identified the hashtag tokens by seeing pound (\#) sign and splitted these based on digits or capital letters. Example: \(\#IndiaFightsCorona \rightarrow India Fights Corona\)

\item \textbf{Stemming: } We removed the inflectional morphemes like ``ed", ``est", ``s", and ``ing" from their token stem. Ex: \(confirmed \rightarrow ``confirm" + ``-ed"\)

\item \textbf{Text cleaning: }To remove the irrelevent data we used this step. Removed punctuation marks, digits and, non-ASCII glyphs from the tweet.
\end{enumerate}

\subsection{Supervised Machine Learning Models}

To build the finest system for fake news detection, we started our investigations with traditional NLP approaches like Linear Regression (LR), Support Vector MAchines (SVM), Passive Agressive Classifier (PAC), XGBoost, and Multi-Layer Perceptron (MLP).  We study the results of above mentioned supervised models with the combination of three types of word vectors:

\begin{enumerate}
    \item Word-level, n-gram level, and character level TF-IDF vectors with the feature matrix size of 100000.
    
    \item English Glove \cite{penn:soch} word embeddings with the dimension of 300.
    
    \item TF-IDF weighted averaging with Glove embeddings. We described below the fake news vector construction. 
\end{enumerate}

\begin{equation}
Tweet_{vector} = \dfrac{\sum\limits_{i=1}^{N} \textbf{tf-idf}(token_i) \times  \textbf{Glove}(token_i)}{\textbf{N}} 
\end{equation}

In the above formula, N is the total number of words in the input fake news tweet, and $token_i$ is the $i^{th}$ token in the input text. After analyzing the results, TF-IDF weighted averaging gave better results than the standard TF-IDF.

\subsection{Deep Learning Models}

Supervised machine learning algorithms performed very well on the provided dataset. In this section, we experiment with deep learning models that give better results than traditional classification algorithms.

\subsubsection{LSTM:}We used Long Short-Term Memory (LSTM) \cite{hoch:schm} architecture with two different pre-trained word embeddings Glove and Fasttext \cite{boja:grav}. LSTM is a type of Recurrent Neural Network (RNN) that can solve long term dependency problem, and it is a well-suited model for sequence classification. 

We converted the input data points into word vectors by using pre-trained word embeddings. These word vectors are passed as input to the LSTM layer. We stacked up two LSTM layers one after another with the dropout of 0.25. The size of LSTM is 128, and the last time step output is treated as input data point representation. The final time step's outcome is passed as an input to a dense layer for fake news detection.
    
\subsubsection{BiLSTM with Attention:}Sometimes not all the tokens in the input text contribute equally to the representation of input text. So we advantage word attention \cite{vasw:shaz} mechanism to catch the tokens' prominent influence on the input data point. We built this attention mechanism on top of BiLSTM layers. 

The sequence of word vector is passed through a BiLSTM layer, which contains one forward and backward LSTM layer. Attention mechanism applied to the output of BiLSTM layer, which produces a dense vector. This dense vector is forwarded to a fully connected network.
    
\subsubsection{CNN:} We explored a Convolution Neural Network (CNN) \cite{lec:beng} model for misinformation detection. The model consists of an embedding layer, a convolution layer with 3 convolutions, a max-pooling layer, and a fully connected network. In the embedding layer, the input texts are converted into $n \times d$ sequence matrix, where n is the length of the input data point and d is the length of the word embedding dimension. In the convolution layer, fed the sequence matrix through three 1D convolutions of kernel sizes 3, 4, and 5. And each convolutions filter size is 128.  The convolution layer's output is max pooled over time and concatenated to get the input datapoint representations in the max-pooling layer. The output of the max-pooling layer is passed to a fully connected network with a softmax output layer.
    
\subsubsection{CNN + BiLSTM:}A CNN and BiLSTM architecture is an ensemble of CNN and bidirectional LSTM models with Fasttext/Glove word embeddings. In this architecture, the CNN extracts the maximum amount of features/information from the input text using convolution layers. The output of CNN becomes the input to BiLSTM, which keeps the data in chronological order in both directions.

The sequence of word vectors are forwarded through a convolution of kernel size 3 with filter size 128. The output of convolution is passed through a BiLSTM. The outcome of BiLSTM is max-pooled over time and followed by one dense layer and a softmax layer.

\subsection{Transformer Models}

This section explored individual and ensembling of the three transformer models BERT, ALBERT, and XLNet. These models have outperformed the other ML and DL algorithms. We implemented these models using HuggingFace\footnote{https://huggingface.co/transformers/} is a PyTorch transformer library. And the hyperparameters of the three models are described in Table 1.

\begin{table}
\begin{center}
\begin{tabular}{c@{\hspace{1pt}} | @{\hspace{1pt}}c@{\hspace{1pt}}|@{\hspace{1pt}} c @{\hspace{1pt}}|@{\hspace{1pt}} c @{\hspace{1pt}} | @{\hspace{1pt}} c @{\hspace{1pt}} | @{\hspace{1pt}} c}
\hline
\textbf{Model} & \textbf{Learning Rate} & \textbf{Batch Size} & \textbf{Optimizer} & \textbf{Max Length} & \textbf{Type} \\
\hline
&&&&&\\
\textbf{BERT}&2e-5  & 16 & Adam & 128 & BERT-Base \\
\textbf{XLNet}&2e-5  & 16 & Adam & 128 & XLNetLarge \\
\textbf{ALBERT}&2e-5  & 32 & Adam & 128 & ALBERT-Xlarge \\
&&&&&\\
\hline
\end{tabular}
\end{center}
\caption{Hyperparameters of transformer models}
\end{table}

\subsubsection{BERT:}Bidirectional Encoder Representations from Transformers (henceforth, BERT) \cite{dev:chan} is a transformer model developed to pre-train deep bidirectional representations from unseen data. This model developed by combining two robust concepts: (i) It's a deep transformer model so that it can process lengthy sentences effectively by using attention mechanism, and (ii) It's a bidirectional network, so it takes into account the entire text passage to comprehend the meaning of each token.

BERT implementation has two steps; one is pre-training and another fine-tuning. In the first step, the model is trained on unseen data over various pre-training problems using a dataset in a particular language or in increases data with multiple languages. In the second step, all the initialized parameters are fine-tuned using the labeled data from certain tasks. 

We fine-tuned the pre-trained BERT (Base) model for our COVID-19 fake news detection task. BERT base model contains the 12 layers of encoder blocks and 12 bidirectional self-attention heads by considering the sequence of 512 tokens and emitting the representations of a sequence of hidden vectors. We added one additional output layer on top of the BERT model to calculate the conditional probability over the output classes, either fake or real.  See FIGURE 1 for the fine-tuned model of BERT.

\begin{center}
  \begin{figure}[h!]
  \makebox[\textwidth]{\includegraphics[width=9cm,height=6cm]{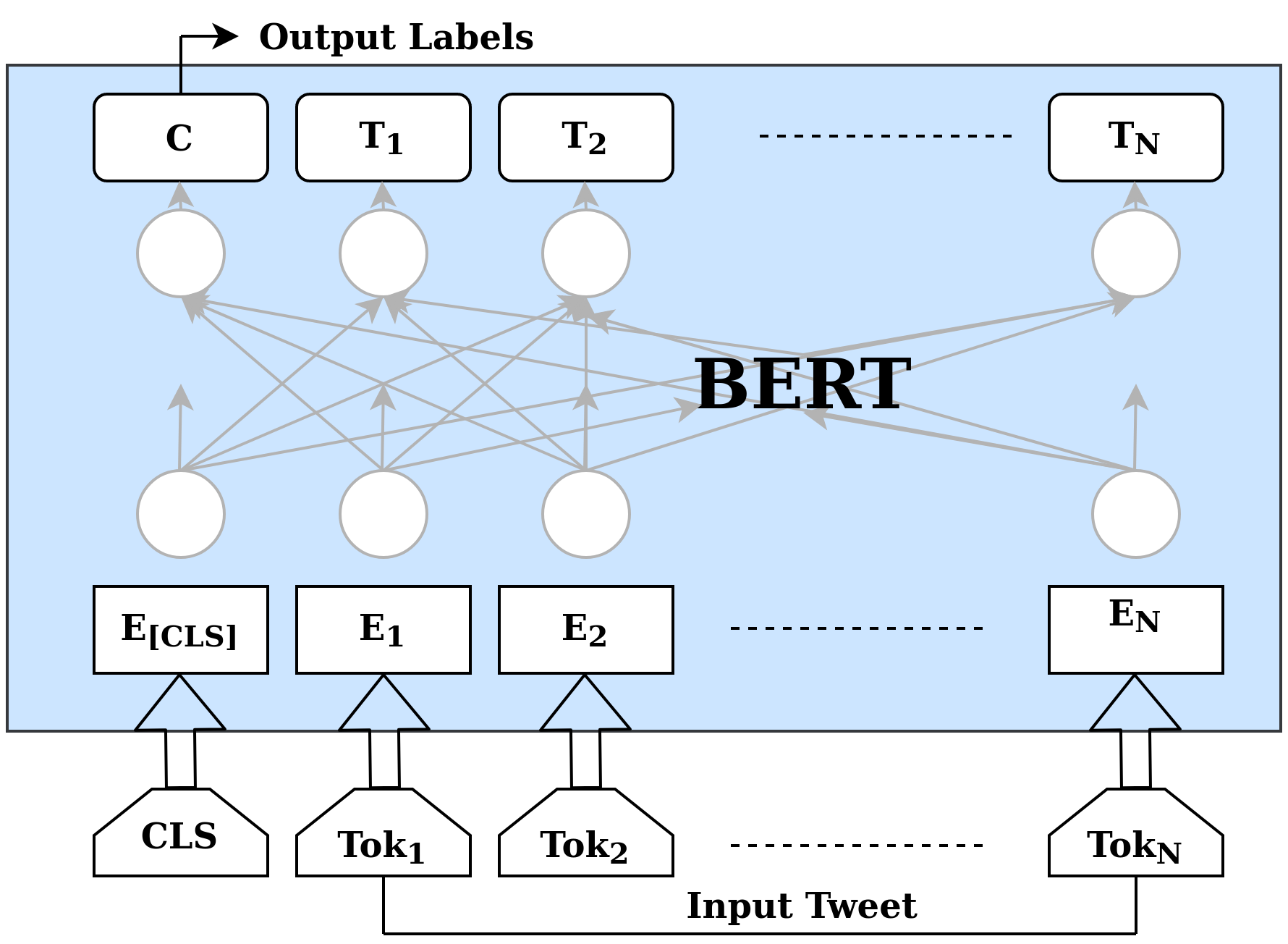}}
  \caption{BERT model architecture}
  \end{figure}
\end{center}

\subsubsection{XLNet:}XLNet is an enhanced version of BERT. To understand the language context deeper, XLNet \cite{yang:dai} uses Transformer-XL \cite{dai:yang} as a feature engineering model, which alone is an adoption upon the native Transformer. This Transformer XL model integrates the two components \textit{Recurrence Mechanism} and \textit{Relative Positional Encoding} (RPE) to the Transformer used in BERT to handle the long-term dependencies for texts that are longer than the maximum allowed input length. Recurrence Mechanism will give context between two sequences at specific segments and RPE, which carries similarity information between two tokens.

The XLNet model has been trained on a huge dataset using the \textit{permutation language modeling}. This technique is one of the main differences between BERT and XLNet, and it uses permutations to generate data from the forward and backward directions at the same time. We used the pre-trained XLNet model from Hugging Face, then fine-tuned the model with a maximum length of 128 to update the pre-trained model to fit our fake news detection dataset.

\subsubsection{ALBERT:} Modern language models increasing the model size and quantity of parameters when pre-training natural language representations. They often give better improvements in many downstream tasks, but in some cases, they become harder due to memory limitation and longer hours of training. To address these problems, a self-supervised learning model ALBERT (A Lite BERT) \cite{lan:chen} often uses parameter reduction techniques to increase model speed and lower memory consumption. We used the A Lite BERT model for our misinformation detection problem, which achieves better performance than DL models. 

\subsubsection{Ensemble Model:}We ensembeled the three transformer models BERT, ALBERT, and XLNet for better prediction. See Figure 4 for the ensemble model. Our ensemble model computes an average of all softmax values from these three transformer models after extracting the softmax probabilities from each model. This model relatively better than other models.   

\begin{center}
  \begin{figure}[h!]
  \makebox[\textwidth]{\includegraphics[width=9cm,height=2.5cm]{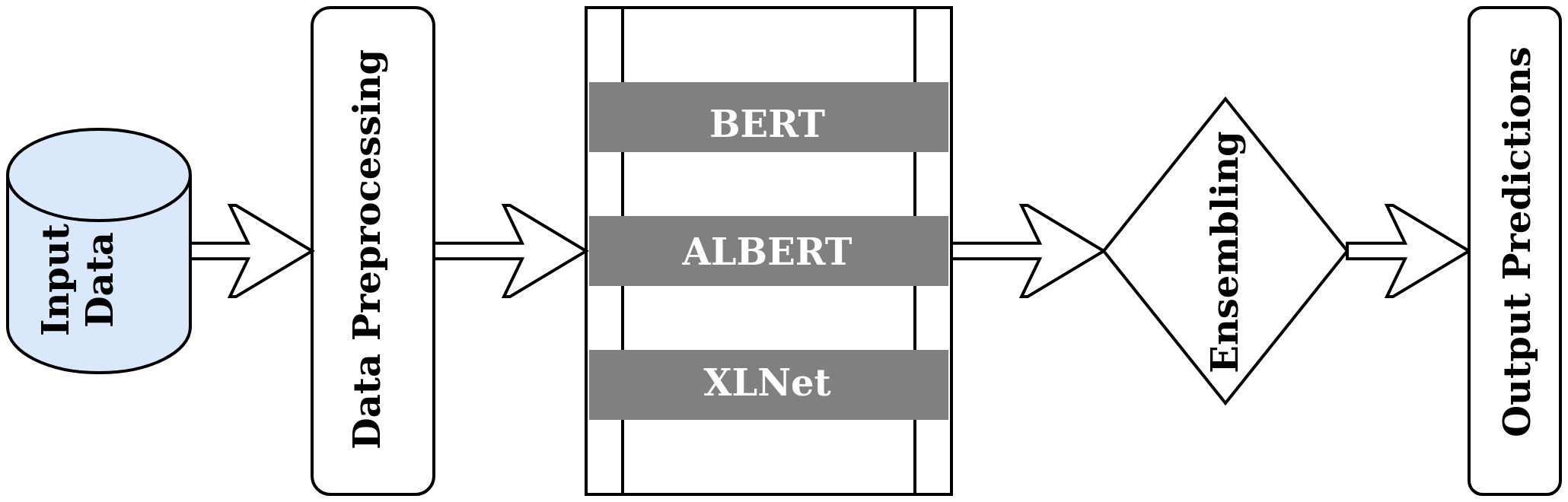}}
  \caption{Transformer based ensemble model architecture}
  \end{figure}
\end{center}

\section{Results and Discussion}

In this section, we compared the performance of various machine learning, deep learning, and transformer-based models using several evaluation metrics like precision, recall, weighted f1-score and accuracy. The results of the various experiments on the test set are reported in Table 3. The results clearly showing that Transformer based models are considerably better than other machine and deep learning models for our COVID-19 misinformation detection task. And while doing experiments, we observed that few models good at retrieving prominent features while other models have the best classification performance.

\begin{table}
\begin{center}
\begin{tabular}{c@{\hspace{0.9pt}} | @{\hspace{0.9pt}}c@{\hspace{0.9pt}}|@{\hspace{0.9pt}} c @{\hspace{0.9pt}}|@{\hspace{0.9pt}} c @{\hspace{0.9pt}} | @{\hspace{0.9pt}} c @{\hspace{0.9pt}} | @{\hspace{0.9pt}} c}
\hline
\textbf{Model Type} & \textbf{Model} & \textbf{Precision} & \textbf{Recall} & \textbf{Accuracy} & \textbf{F1-Score} \\
\hline
&\textbf{SVM}  & 0.9640 & 0.9640 & 0.964013 & 0.964037 \\
\textbf{ML }&\textbf{PAC}  & 0.9673 & 0.9673 & 0.967285 & 0.967289 \\
\textbf{Models}&\textbf{MLP}  & 0.9645 & 0.9645 & 0.964494 & 0.964485 \\
\hline
&\textbf{LSTM with FastText}  & 0.9682 & 0.9682 & 0.9682203 & 0.968224 \\
\textbf{Deep Learning}&\textbf{CNN with FastText}  & 0.9698 & 0.9698 & 0.969802 & 0.969819 \\
\textbf{Models}&\textbf{LSTM + CNN}  & 0.9762 & 0.9762 & 0.976163 & 0.976168 \\
&\textbf{BiLSTM + Attention}  & 0.9790 & 0.9785 & 0.978524 & 0.978504 \\
\hline
&\textbf{BERT}  & 0.9813 & 0.9813 & 0.981306 & 0.981308 \\
\textbf{Transformer}&\textbf{ALBERT}  & 0.9781 & 0.9781 & 0.978031 & 0.978037 \\
\textbf{Models}&\textbf{XLNet}  & 0.9787 & 0.9789 & 0.978596 & 0.978592 \\
&\textbf{Ensemble Model}  & \textbf{0.9855} & \textbf{0.9855} & \textbf{0.985512} & \textbf{0.985514} \\
\hline
\end{tabular}
\end{center}
\caption{Comparision of various fake news detection models on testset}
\end{table}

Classical machine learning models with various TF-IDF feature vectors gave the approximate baseline model results. We observe that the TF-IDF weighted average performed better than the normal TF-IDF vectors. Bi-directional LSTM with attention mechanism f1-score approximate very close to transformer models. The BERT, XLNet, and ALBERT demonstrate better performance than deep learning models. An ensemble of the transformer-based model produces the best F1 score of 0.9855 on the test set. Our transformer based model ranked 5th among 160 teams.

\begin{table}
\begin{center}
\begin{tabular}{c@{\hspace{0.9pt}} |@{\hspace{0.9pt}} c @{\hspace{0.9pt}}|@{\hspace{0.9pt}} c @{\hspace{0.9pt}} | @{\hspace{0.9pt}} c @{\hspace{0.9pt}} | @{\hspace{0.9pt}} c}
\hline
 \textbf{Test Sample} & BERT & ALBERT & XLNet & Ensemble \\
\hline
\#BillGates is shocked that America's pandemic & \cmark & {\xmark} & \xmark & \cmark \\
response is among the worst in the world. &&&&\\
We will all come out stronger from this & \xmark & \xmark & \cmark & \cmark \\
\#COVID \#pandemic. Just \#StaySafeStayHealthy &&&&\\
\hline
\end{tabular}
\end{center}
\caption{Misclassified samples from testset}
\end{table}

In some problems, enesembling of four transformer models is very difficult, and sometimes this approach will not perform well. But if we observe the results of individual transformer models on our dataset are very close, meaning that any transformer model can be used for our fake news detection task. This is the major reason behind the ensembling of transformer models.

In Table 4, we showed the two misclassified test samples. The first test sample actual label is ``real", but only BERT and ensemble models are predicted correctly remaining two models wrongly predicted. And the second sample true label is ``fake", but XLNet and ensemble predicted correctly remaining two models wrongly predicted. However, the ensemble model is correctly predicted in both cases because we are averaging the BERT, ALBERT, and XLNet softmax probabilities. This is a principal observation to ensemble the transformer models.

\section{Conclusion}

In this paper, we presented various algorithms to combat the global infodemic, but transformer-based algorithms performed better than others. And we submitted these models to the Shared Task of COVID-19 fake news detection for English, ConstraintAI-2021 workshop.

Fake news is a progressively significant and tricky problem to solve, particularly in an unanticipated situation like the COVID-19 epidemic. Leveraging state-of-the-art classical and advanced NLP models can help address the problem of COVID-19 fake news detection and other global health emergencies. We intend to explore other contextualized embeddings like FLAIR, ELMo, etc., for a better fake news detecting system in future works.

%
%


\begin{thebibliography}{31}
%
\bibitem{gupta:kumari}
Patwa, P., Bhardwaj, M., Guptha, V., Kumari, G., Sharma, S., PYKL, S., Das, A., Ekbal A., Akhtar, S., Chakraborty.: Overview of CONSTRAINT 2021 Shared Tasks: Detecting English COVID-19 Fake News and Hindi Hostile Posts. (2021). In: Proceedings of the First Workshop on Combating Online Hostile Posts in Regional Languages during Emergency Situation ({CONSTRAINT}). Springer.


\bibitem{datta:yadhav}
Datta, R., Yadav, K., Singh, A., Datta, K., Bansal, A.: The infodemics of COVID-19 amongst healthcare professionals in india. Med. J. Armed
Forces India, vol. 76, no. 3, pp. 276–283, Jul. 2020.

\bibitem{chen:sin}
Chen, X., Sin, S.J.: 'Misinformation? What of it?' Motivations and individual differences in misinformation sharing on social media. In: ASIST (2013)

\bibitem{throne:vla}
Thorne, J., Vlachos, A.: Automated Fact Checking: Task formulations, methods and future directions. In: COLING (2018)

\bibitem{tir:car}
Titcomb, J., Carson, J.: www.telegraph.co.uk. Fake news: What exactly is it – and how can
you spot it?

\bibitem{kali:sing}
Kaliyar, R., Singh, N.: Misinformation Detection on Online Social Media-A Survey. (2019). 1-6. 10.1109/ICCCNT45670.2019.8944587. 

\bibitem{zhang:ghori}
Zhang, X., Ghorbani, A.: An overview of online fake news: Characterization, detection, and discussion. (2020). Inf. Process. Manag., 57, 102025.

\bibitem{khan:khon}
Khan, J.Y., Khondaker, M.T., Iqbal, A., Afroz, S.: A Benchmark Study on Machine Learning Methods for Fake News Detection. (2019). ArXiv, abs/1905.04749.

\bibitem{elha:li}
Elhadad, M., Li, K.F., Gebali, F.: A Novel Approach for Selecting Hybrid Features from Online News Textual Metadata for Fake News Detection. In: Proc. 3PGCIC, Antwerp, Belgium, 2019, pp. 914–925. 

\bibitem{thota:thilak}
Thota, A., Tilak, P., Ahluwalia, S., Lohia, N.: Fake News Detection: A Deep Learning Approach. (2018). SMU Data Science Review: Vol. 1 : No. 3 , Article 10.

\bibitem{sean:doug}
Sean, B., Doug, S., Yuxi, P: Talos Targets Disinformation with Fake News Challenge Victory. (2017). Available online: https://blog.talosintelligence.com/2017/06/talos-fake-news-challenge.html

\bibitem{jwa:oh}
Jwa, H., Oh, D., Park, K., Kang, J., Lim, H.: exBAKE: Automatic Fake News Detection Model Based on Bidirectional Encoder Representations from Transformers (BERT). (2019). Applied Sciences, 9, 4062.

\bibitem{shah:nand}
Shahi, G.K., Nandini, D.: FakeCovid - A Multilingual Cross-domain Fact Check News Dataset for COVID-19. (2020). ArXiv, abs/2006.11343.

\bibitem{zhou:mulay}
Zhou, X., Mulay, A., Ferrara, E., Zafarani, R.: ReCOVery: A Multimodal Repository for COVID-19 News Credibility Research. (2020). In: Proceedings of the 29th ACM International Conference on Information \& Knowledge Management.

\bibitem{cui:lee}
Cui, L., Lee, D.: CoAID: COVID-19 Healthcare Misinformation Dataset. (2020). ArXiv, abs/2006.00885.

\bibitem{memon:car}
Memon, S.A., Carley, K.M.: Characterizing COVID-19 Misinformation Communities Using a Novel Twitter Dataset. (2020). ArXiv, abs/2008.00791.

\bibitem{li:jiang}
Li, Y., Jiang, B., Shu, K., Liu, H.: MM-COVID: A Multilingual and Multimodal Data Repository for Combating COVID-19 Disinformation. (2020). ArXiv, abs/2011.04088.

\bibitem{rakh:amri}
Al-Rakhami, M.S., Al-Amri, A.M.: Lies Kill, Facts Save: Detecting COVID-19 Misinformation in Twitter. (2020). IEEE Access, 8, 155961-155970.

\bibitem{vijj:potl}
Vijjali, R., Potluri, P., Kumar, S., Teki, S.: Two Stage Transformer Model for COVID-19 Fake News Detection and Fact Checking. (2020).

\bibitem{hoss:robe}
Hossain, T., RobertL.Logan, I., Ugarte, A., Matsubara, Y., Young, S.,Singh, S.: COVIDLies: Detecting COVID-19 Misinformation on Social Media. (2020). NLP4COVID@EMNLP.

\bibitem{patwa:sharma}
Patwa, P., Sharma, S., Pykl, S., Guptha, V., Kumari, G., Akhtar, M.S., Ekbal, A., Das, A., Chakraborty, T. (2020). Fighting an Infodemic: COVID-19 Fake News Dataset. ArXiv, abs/2011.03327.


\bibitem{elha:lik}
Elhadad, M.K., Li, K., Gebali, F.: Detecting Misleading Information on COVID-19. (2020). IEEE Access, 8, 165201-165215.

\bibitem{penn:soch}
Pennington, J., Socher, R., Manning, C.D.: Glove: Global Vectors for Word Representation. (2014). In: EMNLP.

\bibitem{hoch:schm}
Hochreiter, S., Schmidhuber, J.: Long Short-Term Memory. (1997). Neural Computation, 9, 1735-1780.

\bibitem{boja:grav}
Bojanowski, P., Grave, E., Joulin, A., Mikolov, T.: Enriching Word Vectors with Subword Information. (2017). Transactions of the Association for Computational Linguistics, 5, 135-146.

\bibitem{vasw:shaz}
Vaswani, A., Shazeer, N., Parmar, N., Uszkoreit, J., Jones, L., Gomez, A.N., Kaiser, L., Polosukhin, I.: Attention is All you Need. (2017). NIPS.

\bibitem{lec:beng}
LeCun, Y., Bengio, Y., Hinton, G.: Deep learning. (2015). Nature, 521(7553), pp.436-444.

\bibitem{dev:chan}
Devlin, J., Chang, M., Lee, K., Toutanova, K.: BERT: Pre-training of Deep Bidirectional Transformers for Language Understanding.  (2019). NAACL-HLT.

\bibitem{yang:dai}
Yang, Z., Dai, Z., Yang, Y., Carbonell, J., Salakhutdinov, R., Le, Q.V.: XLNet: Generalized Autoregressive Pretraining for Language Understanding. (2019). NeurIPS.

\bibitem{dai:yang}
Dai, Z., Yang, Z., Yang, Y., Carbonell, J., Le, Q.V., Salakhutdinov, R.: Transformer-XL: Attentive Language Models Beyond a Fixed-Length Context. (2019). ACL.

\bibitem{lan:chen}
Lan, Z., Chen, M., Goodman, S., Gimpel, K., Sharma, P., Soricut, R.: ALBERT: A Lite BERT for Self-supervised Learning of Language Representations. (2020). ArXiv, abs/1909.11942.

\end{thebibliography}
\end{document}